%% file: bare_jrnl.tex
\documentclass[journal]{IEEEtran}
\ifCLASSINFOpdf
\else
\fi
\usepackage{url}
\usepackage{booktabs}
\usepackage{multirow}

\usepackage{cite}
\usepackage{amsmath,amssymb,amsfonts}
\usepackage{algorithmic}
\usepackage{graphicx}
\usepackage{textcomp}
\usepackage{colortbl}
\usepackage{xcolor}
\usepackage{array}
\usepackage{multirow}
\begin{document}
\newcommand{\etal}{\textit{et al}. }
\newcommand{\ie}{\textit{i}.\textit{e}., }
\newcommand{\eg}{\textit{e}.\textit{g}. }
\title{Generalized Out-of-distribution Fault Diagnosis
(GOOFD) via Internal Contrastive Learning}
\author{Xingyue Wang$^\dagger$,
        Hanrong Zhang$^\dagger$, Xinlong Qiao, Ke Ma, Shuting Tao, Peng Peng$^\ast$,~\IEEEmembership{Member,~IEEE}, Hongwei Wang$^\ast$,~\IEEEmembership{Member,~IEEE}
% \thanks{This paragraph of the first footnote will contain the date on 
% which you submitted your paper for review. It will also contain support 
% information, including sponsor and financial support acknowledgment. For 
% example, ``This work was supported in part by the U.S. Department of 
% Commerce under Grant BS123456.'' }
\thanks{$^\dagger$Equal contribution.}
\thanks{Manuscript received 30 June 2023; revised 12 November 2023 and 12 February 2024; accepted 4 April 2024. Date of publication xxxx; date of current version 4 April 2024. This work was supported in part by the National Natural Science Foundation of China under Grant 62276230. Paper no. TII-23-4488. (Corresponding authors: Hongwei Wang and Peng Peng.)}
\thanks{Xinyue Wang, Hanrong Zhang, Xinlong Qiao, Peng Peng, and Hongwei Wang are with Zhejiang University and the University of Illinois Urbana–Champaign Joint Institute, Haining, 314400, China. (e-mail: xinyue1.22@intl.zju.edu.cn, hanrong.22@intl.zju.edu.cn, xinlong.23@intl.zju.edu.cn, pengpeng@intl.zju.edu.cn, hongweiwang@intl.zju.edu.cn)}
\thanks{Ke Ma, Shuting Tao are with the College of Computer Science and Technology in Zhejiang University, Hangzhou, 310013, China. (e-mail: ke.22@intl.zju.edu.cn, shuting.17@intl.zju.edu.cn).}}

\markboth{IEEE Transactions on Industrial Informatics}%
{Shell \MakeLowercase{\textit{et al.}}: Enhance Fault Diagnosis Performance By Neural Network Ari}



\maketitle

\begin{abstract}
Fault diagnosis is crucial in monitoring machines within industrial processes. With the increasing complexity of working conditions and demand for safety during production, diverse diagnosis methods are required, and an integrated fault diagnosis system capable of handling multiple tasks is highly desired. However, the diagnosis subtasks are often studied separately, and the current methods still need improvement for such a generalized system. To address this issue, we propose the Generalized Out-of-distribution Fault Diagnosis (GOOFD) framework to integrate diagnosis subtasks. Additionally, a unified fault diagnosis method based on internal contrastive learning and Mahalanobis distance is put forward to underpin the proposed generalized framework. The method involves feature extraction through internal contrastive learning and outlier recognition based on the Mahalanobis distance. Our proposed method can be applied to multiple faults diagnosis tasks and achieve better performance than the existing single-task methods. Experiments are conducted on benchmark and practical process datasets, indicating the effectiveness of the proposed framework.

\end{abstract}


% \fi
\IEEEpeerreviewmaketitle

\input{main.tex}

% or



\ifCLASSOPTIONcaptionsoff
  \newpage
\fi

\end{document}

%% file: main.tex
\section{Introduction}
\label{sec:introduction}
\IEEEPARstart{D}{uring} the industrial process and mechanical operation, faults inevitably occur, resulting in declining industrial efficiency and severe losses~\cite{10114639,10152774,li2023sccam}. Accurate fault diagnosis has become increasingly crucial in system design and maintenance to ensure safety in machinery production~\cite{8680674,9072621,s22218537} using the burgeoning deep learning technology~\cite{ li2023order,zhang2023saka,wang2023hard, wang2023weighted,wang2023imbalanced}. 

Currently, the research in fault diagnosis mainly contains two tasks: 1) determining whether the system is normal. The technique for this task is called process monitoring, which aims to detect the abnormal during production. 2) distinguishing \text{blue}{faults that occurred} while finding their reasons. The methods for this task are fault detection and fault classification. 
In process monitoring, the traditional approaches applied the multivariate statistical method, such as Principal Components Analysis (PCA)~\cite{7795255}, Partial Least Squares (PLS)~\cite{abdi2013partial}, and Independent Component Analysis (ICA)~\cite{GE20131}, which extract features by transforming the raw data into a lower dimension and detecting occurred faults simply by statistical variance. 
In recent years, some new tasks have emerged to deal with more complex production situations, such as a novel task called open set fault diagnosis (OSFD), which detects unseen faults that occur during industrial processes. It is a relatively new task in fault diagnosis, aiming to detect unknown fault samples while correctly classifying known classes. Transfer Learning (TL) and domain adaptation (DA) combined with other deep-learning methods have been widely utilized in the OSFD task~\cite{9905947,9994041}. Chen et al. \cite{9863762} proposed a multi-source open-set DA diagnosis approach to tackle the OSFD task. Mao et al. \cite{MAO2022111125} introduced an interactive dual adversarial neural network (IDANN), which uses weighted unknown classification items to distinguish the outliers. Li et al. \cite{9005182} established a deep adversarial transfer learning network (DATLN), where a classifier is trained to learn the decision boundary and detect new faults. 
The tasks and techniques mentioned above are often discussed and tested separately. Nevertheless, since practical production is complicated, it leads to two problems: 
1) A generalized fault diagnostic framework for multiple tasks is still lacking
2) The training on different techniques is time-consuming since they have different feature extraction methods and network structures. 

To address the problems, it is crucial to develop a generalized fault diagnosis system that considers several situations with a unified method to deal with multiple tasks. 
It enables comprehensive monitoring of a machine's status, which facilitates the timely detection and identification of any faults that may arise, allowing for prompt action to be taken.
Moreover, the process monitoring and OSFD methods mentioned above still need to be improved for the generalized framework. First, these methods can only solve a single task. They cannot simultaneously deal with all subtasks in a generalized framework, which does not match the comprehensive approach we need. Secondly, the working environment changes during the practical production process, and the above methods do not perform well in detecting faults under variable condition scenarios due to the lack of a powerful feature extractor. As a result, internal contrastive learning (ICL), as an effective representation learning technique, can be consequently used in our method.

In recent years, contrastive learning has been used to extract features from data \cite{10196057,peng2023sclifdsupervisedcontrastiveknowledgedistillation}. It is a stochastic data augmentation method in self-supervised learning, aiming to pull similar samples closer while pushing different samples apart in the embedding space. This method has proven effective in feature representation learning \cite{qiu2021neural} in numerous research studies. Currently, some studies have applied contrastive learning to fault diagnosis methods to extract features and improve performance under changing working conditions. The contrastive-learning-based diagnosis methods are able to learn invariant features and improve performance under changing operating conditions.
Zhang \cite{10018491} et al. introduced a semi-supervised contrastive learning method to enhance the feature mapping of limited data.  
Li \cite{10042974} et al. used contrastive learning for the fault diagnosis of rolling element bearings to extract interclass features. Zhang et al. \cite{10032199} proposed a contrastive learning-based diagnosis model to learn domain-invariant fault features and improve performance under variable working conditions. 
In recent research, Tom and Wolf \cite{shenkar2021anomaly} have utilized ICL for anomaly detection. This method uses subvectors from given samples as positive/negative pairs to generate the contrastive loss, which depends little on the data structure and can better detect outliers.

Inspired by the above, this paper proposes a novel framework, and the main contributions are summarized as follows:

\begin{itemize}

 \item We are the first to propose an integrated diagnostic system called the Generalized Out-of-distribution Fault Diagnosis (GOOFD) Framework, including process monitoring, fault classification, and OSFD tasks, which offers significant room for expansion and exploration.

\item The paper introduces a novel integrated diagnosis method ICL-OD to 1) address the multi-tasks issue in the GOOFD framework and 2) learn more distinctive features for unknown classes based on internal contrastive learning and the Mahalanobis-distance approach.

\item Extensive experiments on benchmark datasets and actual mechanical datasets are conducted to indicate that our proposed method can be applied to different fault diagnosis tasks and has a better performance compared to the existing single-task methods.

\end{itemize}

The rest of the paper is organized as follows: Section II presents the motivation and background of our proposed framework and methods. Section III introduces the integrated fault diagnosis framework and the proposed methods. Section IV presents the experimental results and the validation of the proposed model. Finally, the conclusion of the study is presented in Section V.

\section{Background and Motivation}
\subsection{Motivation}
There are increasing numbers of new tasks and demands for fault diagnosis in industrial production, and many of these tasks have numerous subtasks that require various techniques. In the production process, there are often situations where many different types of faults need to be diagnosed, such as fault detection, fault classification, and unseen fault detection, which cannot be achieved through a single task alone. A fault diagnosis system that can comprehensively solve multiple tasks is required to diagnose faults more efficiently.

Currently, subtasks are discussed separately and have yet to be integrated to implement a comprehensive fault diagnosis system, and specific steps in the tasks, such as feature extraction and fault recognition, are also carried out separately. Thus, it is essential to build a comprehensive diagnostic system. We have found that the diagnosis subtasks have some commonalities, such as all involving the detection of outliers, indicating that it is feasible to combine these subtasks to solve them, and integrating them can also improve the efficiency of model training by using a unified feature map. Therefore, we propose this integrated framework to consolidate these subtasks.

Our method treats the unified problem in the proposed framework as the outlier detection task. In process monitoring, fault samples are defined as the outliners. In OSFD, the outliners are the unknown fault samples. This way, the generalized fault diagnosis issue can be treated as the out-of-distribution detection problem, in which the outliers vary from specific subtasks. Therefore, we used a contrastive learning method that is more suitable for outlier detection: ICL. The ICL technique generates positive and negative samples from the internal sub-vectors of the given sample, which focuses more on learning its characteristics and is more suitable for outlier detection problems in binary classification. By integrating ICL into our method, the detection of outliers is improved, achieving better fault detection performance.

\subsection{Contrastive Learning}

Contrastive learning is a stochastic data augmentation method in self-supervised learning, which is proven effective in feature representation learning \cite{qiu2021neural}. 
For a batch of $N$ samples, the model generates $2N$ augmented data, where each source data $x$ has two augmentations $\tilde{x}_{i}$ and $\tilde{x}_{j}$. For each augmented data $\tilde{x}_{i}$, the other augmented data $\tilde{x}_{j}$ generated from the same source data $x$ is regarded as the positive pair. The remaining $2(N-1)$ augmentations $\tilde{x}_{j}(m \neq j)$ generated from other $(N-1)$ source data is deemed as the negative pair of $\tilde{x}_{i}$. The model is trained to minimize the distance between every positive pair and maximize the distance between every negative pair. The contrastive loss is defined as:

\begin{equation}
\mathcal{L}= -\sum_{i\in A}log{\frac{exp(f(\tilde{x}_{i})\cdot f(\tilde{x}_{j})/\tau)}{\sum_{m \neq j} exp(f(\tilde{x}_{i})\cdot f(\tilde{x}_{m})/\tau)}},\label{contrastiveloss}
\end{equation}
where $f(\tilde{x}_{k})$ represents the embedding of augmented data $\tilde{x}_{k}$ by encoder $f(\cdot)$. The index $i \in A$ ranges from $(0, 2N)$.

\subsection{Internal Contrastive Learning}
In \cite{shenkar2021anomaly}, Tom and Wolf proposed an anomaly detection method based on ICL. The method sliced the input vector into different dimensions. The network is trained to produce similar embeddings for complementary pairs and push away every sub-vector pair in the input vector. Based on the internal contrastive loss, the method generates positive and negative pairs from the subvectors of a sample ${x}_{i}$ with dimension $D$ to learn the representations of each sample. To split the vectors, here, we set a hyperparameter $d$ to represent the starting index of a sub-vector, also called the \emph{internal dimension} of the subset pairs. First, given the parameter $d$, extract consecutive $l$-length sub-vector ${p}_{i}^{d} =\{{x}_{i}^{d},...,{x}_{i}^{d+l-1}\}$ from ${x}_{i}$. Then, the complementary part ${q}_{i}^{d} =\{{x}_{i}^{1},...,{x}_{i}^{d-1},{x}_{i}^{d+l},...,{x}_{i}^{D}\}$ with the length of $D-l$, is defined as the positive pair of ${p}_{i}^{d}$. For index $d' \neq d$, ${p}_{i}^{d'}$ is regarded as the negative pair of ${q}_{i}^{d}$. The method has two encoders $F$ and $G$ to learn the feature map of ${p}_{i}^{d}$ and ${q}_{i}^{d}$, respectively. Here, we normalized the networks and obtained the normalized networks ${F}^{N}$ and ${G}^{N}$. Let $\mathcal{D}({q}_{i}^{d},{p}_{i}^{d})$ represent the dot product between ${p}_{i}^{d}$ and ${q}_{i}^{d}$, which is shown in \eqref{distance} :

\begin{equation}
    \mathcal{D}({q}_{i}^{d},{p}_{i}^{d})={F}^{N}({q}_{i}^{d})\cdot{G}^{N}({p}_{i}^{d}).
    \label{distance}
\end{equation}

The two networks are trained to learn similar embeddings between positive pairs ${F}^{N}({p}_{i}^{d})$ and ${G}^{N}({q}_{i}^{d})$ while minimizing the mutual information between negative pairs ${p}_{i}^{d'}$ and ${q}_{i}^{d}$. The internal contrastive loss for sample ${x}_{i}$ in internal dimension $d$ is defined in \eqref{InternalContrastiveLoss}, where $k = D-l+1$: 

\begin{equation}
    \ell({x}_{i},j)=-ln\frac{{e}^{\mathcal{D}({q}_{i}^{d},{p}_{i}^{d})/\tau}}{\sum_{d'\neq d}^{k}{e}^{\mathcal{D}({q}_{i}^{d},{p}_{i}^{d'})/\tau}}.
    \label{InternalContrastiveLoss}
\end{equation}

In the internal contrastive loss, for a given sample 
${x}_{i}$ and internal dimension $d$, the method learns to produce similar embeddings ${q}_{i}^{d}$ and ${p}_{i}^{d}$ between the matched sub-vectors from the same dimension $j$ and separate the $l$-length sub-vectors ${q}_{i}^{d}$ and ${p}_{i}^{d'}$ from different dimensions. Then the model can generate scores $S({x}_{i})$ for ${x}_{i}$ considering all the internal dimensions $d=\{0,..., k\}$, as shown in \eqref{score}:

\begin{equation}
    S({x}_{i})={\textstyle \sum_{d}\ell({x}_{i}, d)}, 
    \label{score}
\end{equation}

where the final output score $S({x}_{i})$ denotes the representation of the sample ${x}_{i}$.

\section{Generalized Out-of-distribution Fault Diagnosis}
\begin{figure*}[t]
    \centerline{\includegraphics[width=\textwidth]{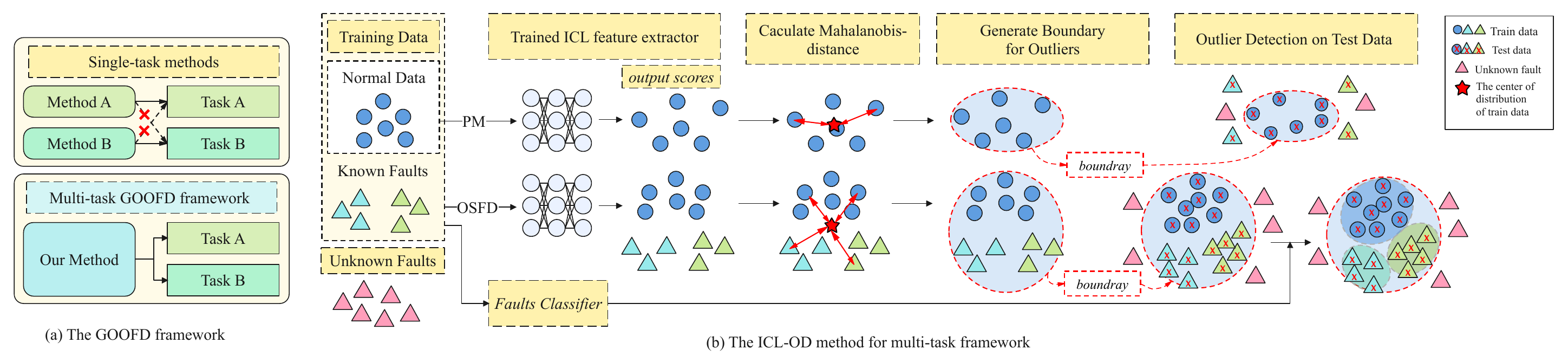}}
    \caption{(a) The difference between the GOOFD framework and existing single-task methods. A single-task approach can only handle a specific task. However, the GOOFD framework is dedicated to integrating and solving various fault diagnosis tasks through a unified method. (b) The basic process of our approach addresses multi-task issues.}
    \label{framework}
\end{figure*}

The framework of the proposed GOOFD system is shown in Fig.\ref{framework}. The framework uses the ICL-OD method to learn the boundary for outliers and then detect outliers to complete both process monitoring and OSFD tasks. The following subsections first introduce the definition of the unified tasks in the GOOFD system and then describe the methodology of the ICL-OD approach and how it is used to tackle multiple tasks in the GOOFD system.

\subsection{Task Definition}
Given a set of $n$ normal data $\mathcal{D}_{0}=\{ {x}_{1}, ...,{x}_{n} \}$, categoried in the normal class ${X}_{0}$, and a set of known fault data $\mathcal{D}_{f}=\{ {f}_{1}, ...,{f}_{m} \}$, which belong to $(N-1)$ fault classes ${X}_{1},...,{X}_{N-1}$. The embedding space for normal data $\mathcal{D}_{0}$ is called ${\mathcal{S}}_{0}$ and the embedding space for the known fault data $\mathcal{D}_{f}$ is called ${\mathcal{S}}_{f}$. For the unknown fault $\mathcal{D}_{u}=\{ {u}_{1}, ...,{u}_{k} \}$ with embedding space ${\mathcal{S}}_{u}$, the label of ${u}_{i}$ is class ${X}_{N}$. The embedding space of known data is defined as ${\mathcal{S}}_{k}$. Its corresponding open space is defined as ${\mathcal{O}}_{k}$. 

In the proposed GOOFD, The embedding space ${\mathcal{S}}_{0}$, ${\mathcal{S}}_{f}$, and ${\mathcal{S}}_{u}$ collectively form the entire embedding space. The process monitor and OSFD tasks are fused by specifying different boundaries for the known embedding space $\mathcal{S}_{k}$.
In process monitoring, the known embedding space contains only normal data $\mathcal{D}_{0}$. Thus it is defined as ${\mathcal{S}}_{k} = \mathcal{S}_{0}$. Its corresponding open space $\mathcal{O}_{0}$ is $\mathcal{S}_{f} \cup \mathcal{S}_{u}$, which contains all the faults. While in OSFD, the known embedding space ${\mathcal{S}}_{k}$ consists of normal data $\mathcal{D}_{0}$ and known faults $\mathcal{D}_{f}$, and its corresponding ${\mathcal{O}}_{k} = \mathcal{S}_{u} $.

\subsection{ICL for Outlier Detection}
The proposed ICL for Outlier Detection (ICL-OD) method can be divided into two steps:  generate the embedding map based on ICL and detect the outliers. The ICL-OD applied the ICL-based method for fault diagnosis in \cite{shenkar2021anomaly} to extract features from the input samples. Given a set of input samples $X=\{ {x}_{1}, ...,{x}_{n} \}$, obtain score $S({x}_{i})$ as the feature space for each sample ${x}_{i}$ based on \eqref{score}.

After obtaining the score $S({x}_{i})$, the method uses a Mahalanobis-distance-based classifier to detect the outlier. The Mahalanobis distance measures the distance between a sample and a distribution. Given an input sample $x$ and a distribution $Q$, the  Mahalanobis distance ${d}_{M}(x)$ between them is defined in \eqref{md}: 
\begin{equation}
    {d}_{M}(x)=\sqrt{(x-\mu){\sum}^{-1}(x-\mu)},
    \label{md}
\end{equation}
where $\mu$ represents the means of $Q$ and $\sum$ represents the covariance matrix of $Q$. Unlike Euclidean distance, the Mahalanobis distance considers associations between correlated variables. It has been proven to be an effective graphic tool for identifying outliers. In our method, the Mahalanobis distance can be applied to measure the distance ${d}_{S}(x)$ between sample $x$ and the distribution of the training set (normal data) in the score embedding space, as shown in \eqref{mds}: 
\begin{equation}
    {d}_{S}(x)=\sqrt{(S(x)-{\mu}_{S}){\sum}_{S}^{-1}(S(x)-{\mu}_{S})},
    \label{mds}
\end{equation}
where $S(x)$ denotes the output score of sample $x$,  ${\mu}_{S}$ and ${\sum}_{S}$ represent the means and covariance matrix of score $S$ for training sets, respectively. In the one-classification task, the training set contains only normal data, so the output score $S$ can be deemed as the distribution of normal data. 

Given training samples, the method first implements the ICL method to obtain the output score $S$ for training input samples and calculate the means ${\mu}_{tr}$ and covariance matrix ${\sum}_{tr}$ of $S$. Then, for every sample $\{{x}_{i}^{tr}| i\in (1,n)\}$ in the training set, calculate the Mahalanobis distance ${d}_{S}^{train}=\{{d}_{S}({x}_{i}^{tr})| i\in (1,n)\}$ of each training sample to the normal distribution, as shown in \eqref{mdtrain}:
\begin{equation}
{d}_{S}({x}_{i}^{tr})=\sqrt{(S({x}_{i}^{tr})-{\mu}_{tr}){\sum}_{tr}^{-1}(S({x}_{i}^{tr})-{\mu}_{tr})}.
    \label{mdtrain}
\end{equation}

Similarly, for every sample $\{{x}_{i}^{te}|i\in (1,m)\}$ in the testing set, calculate the Mahalanobis distance ${d}_{S}^{test}=\{{d}_{S}({x}_{i}^{te})|i\in (1,m)\}$ of each test sample to the normal distribution. Based on the distance, the outliers can be identified, as shown in \eqref{detection}:

\begin{equation}
    {y}=  \begin{array}{lr} 
  \left\{\begin{matrix} 
  \text{\emph{inliner}}, & {d}_{S}({x}^{te})\le {\theta}, \\ 
  \text{\emph{outlier}}, & {d}_{S}({x}^{te})>{\theta},
\end{matrix}\right. \end{array} 
    \label{detection}
\end{equation}
where threshold $\theta$ is used to reject outliers, and its value is equal to the $k$\% quantile of Mahalanobis distance in ${d}_{S}^{train}$. If ${d}_{S}({x}_{i}^{te})$ exceeds threshold $\theta$, the sample ${x}_{i}^{te}$ will be determined as the outlier, and vice versa, is determined as the normal.

\subsection{Generalized Out-of-distribution Detection}
As previously described, the GOOFD fault diagnosis task contains both process monitoring and OSFD tasks for fault diagnosis. The ICL Outlier Detection (ICL-OD) method proposed in this paper can be used in GOOFD fault diagnosis for process monitoring and OSFD, as shown in Figure 1. Since the process monitoring task aims at detecting the fault from normal, the input data $x$ only consists of normal data of class ${X}_{0}$. Then, the ICL-OD method helps to generate the score $S$ for normal data and reject the anomaly based on $d_{S}({x})$, as shown in \eqref{ad1}:
\begin{equation}
    d_{S}({x})=\sqrt{(S({x})-{\mu}_{0}){\sum}_{0}^{-1}(S({x})-{\mu}_{0})},
    \label{ad1}
\end{equation}
where ${\mu}_{0}$ and ${\sum}_{0}$ represent the means and covariance matrix of score $S$ for normal samples. Then, use the $k$\% quantile of Mahalanobis distance of normal samples as the threshold ${\theta}_{0}$ to find the outliers, as shown in \eqref{ad2}:

\begin{equation}
{y}=  \begin{array}{lr} 
  \left\{\begin{matrix} 
  0, & {d}_{S}(x)\le {\theta}_{0}, \\ 
  1, & {d}_{S}(x)>{\theta}_{0}.
\end{matrix}\right. \end{array} 
\label{ad2}
\end{equation}

For a sample $x$ in the test set, if the Mahalanobis distance ${d}_{S}(x)$ of $x$ exceeds the threshold ${\theta}_{0}$, the sample $x$ is labeled as 1 and rejected as the outliers. Otherwise, it will be labeled as 0 and classified as normal.

In OSFD, which is a multi-class classification task, the training data contains normal data and known faults, regarded as known classes. First, the method trains a classifier $\mathcal{C}$ to classify known classes, which include the normal data class ${X}_{0}$ and $N-1$ known fault class ${X}_{1},...{X}_{N}$. The input samples are regarded as the known classes in OSFD. Then, the ICL-OD method is used to learn the boundary of the empirical data in Mahalanobis distance. Similar to Anomaly Detection, for a given sample $x$ from the testing set, the method uses the Mahalanobis-distance-based threshold to reject the unknown, shown in \eqref{open-set fault diagnosis1}:
\begin{equation}
    d_{S}({x})=\sqrt{(S({x})-{\mu}_{k}){\sum}_{k}^{-1}(S({x})-{\mu}_{k})},
    \label{open-set fault diagnosis1}
\end{equation}
% The ${x}_{i}$ represents data in known class ${X}_{0}$ and ${X}_{1},...{X}_{N}$. 
where ${\mu}_{k}$ and ${\sum}_{k}$ denote the means and covariance matrix of the known training samples, respectively. The method uses the $k$\% quantile of Mahalanobis distance of training samples as the threshold ${\theta}_{u}$ to reject the unknown.
Finally, the method can distinguish between known and unknown, shown in \eqref{open-set fault diagnosis2}:
\begin{equation}
{y}=  \begin{array}{lr} 
  \left\{\begin{matrix} 
  argmax(\mathcal{C}({x}))\in(1, N), & {d}_{S}(x)\le {\theta}_{u}, \\ 
  N+1, & {d}_{S}(x)>{\theta}_{u}.
\end{matrix}\right. \end{array} 
\label{open-set fault diagnosis2}
\end{equation}

The trained classifier $\mathcal{C}$ firstly outputs the softmax score of ${x}$ and classifies them into known classes $1\sim N$. Then, for all the samples whose Mahalanobis distance is greater than the threshold ${\theta}_{u}$, the method will treat these points as outliers and classify them into the unknown class (class $N+1$).

\section{Experiment Study}
In this section, we evaluate and compare our proposed method with baseline methods and set up two experiments to prove the effectiveness of the proposed method. In Experiment I, we first use a simulated benchmark dataset to perform a preliminary validation of the proposed method. In Experiment II, the evaluation of our approach is further verified using two datasets generated from the real-world process. The details of the two experiments are listed below.

\begin{enumerate}
\item[-]Experiment I: The proposed method will be evaluated in the Tennessee Eastman process \cite{LAWRENCERICKER1996205} to measure the performance in process monitoring and OSFD for fault diagnosis. In process monitoring, the dataset uses normal data as the training set and fault data as the test set. The OSFD task divides the dataset into open and closed sets to test its ability to detect unseen faults. 

\item[-]Experiment II: To further evaluate the effectiveness of our method, Experiment II implements the proposed technique with a real-world dataset: Multiphase Flow Facility (MFF) Dataset \cite{MFF}. The MFF dataset is used in process monitoring and OSFD tasks, respectively.
\end{enumerate}

\noindent
\textbf{Baseline.} In Experiment I, for the process monitoring task, we compare the proposed method with the following seven baseline methods: PCA \cite{7795255}, Kernel PCA (KPCA) \cite{7795255}, ICA \cite{GE20131}, Kernel ICA (KICA) \cite{KICAresults}, Dynamic PCA (DPCA) \cite{RATO2013101}, Structured Joint Sparse Canonical Correlation Analysis (SJCCA) \cite{9068308}, Orthonormal Subspace Analysis (OSA) \cite{LOU2022110148}, and Decentralized Support Vector Data Descriptions (DSVDD) \cite{wang2023decentralized}. 
In the OSFD task, we compare the proposed method with the following six baselines:  SoftMax, OpenMax \cite{OpenMax}, CenterLoss \cite{wen2016discriminative}, EOW-SoftMax \cite{EOW-Softmax}, Generalized ENtropy score (GEN) \cite{xixi2023GEN}, and Positive and Unlabeled learning and Label Shift Estimation (PULSE) \cite{garg2022domain}.
In addition, we compare our proposed method with the ICL anomaly detection approach \cite{shenkar2021anomaly} to validate the superiority of our method. Since the ICL anomaly detection approach cannot be directly applied to the OSFD task, we add a classifier to the method to complete the experiment, which we call \textbf{ICL+} in the later section.

Experiment II implements PCA, DPCA, PLS, Enhanced Dynamic Latent Variable (EDLV) analysis \cite{WANG2024105292}, and Global–local Preserving Projection based on Optimal Active Relative Entropy (OARE-GLPP) \cite{liu2022industrial} as the baseline methods in the process monitoring task. The baseline used in the OSFD task remains SoftMax, OpenMax, CenterLoss,  EOW-Softmax, GEN, PULSE, and ICL+.

\noindent
\textbf{Metrics.} In process monitoring, our experiments use the Fault Detection Rate (FDR) to evaluate a model's ability to detect faults. In OSFD, we use the F1 score and AUROC to assess performance. 

\subsection{Experiment I: evaluation with TE Process}

\begin{figure}[t]
    \centering
    \includegraphics[width=\columnwidth]{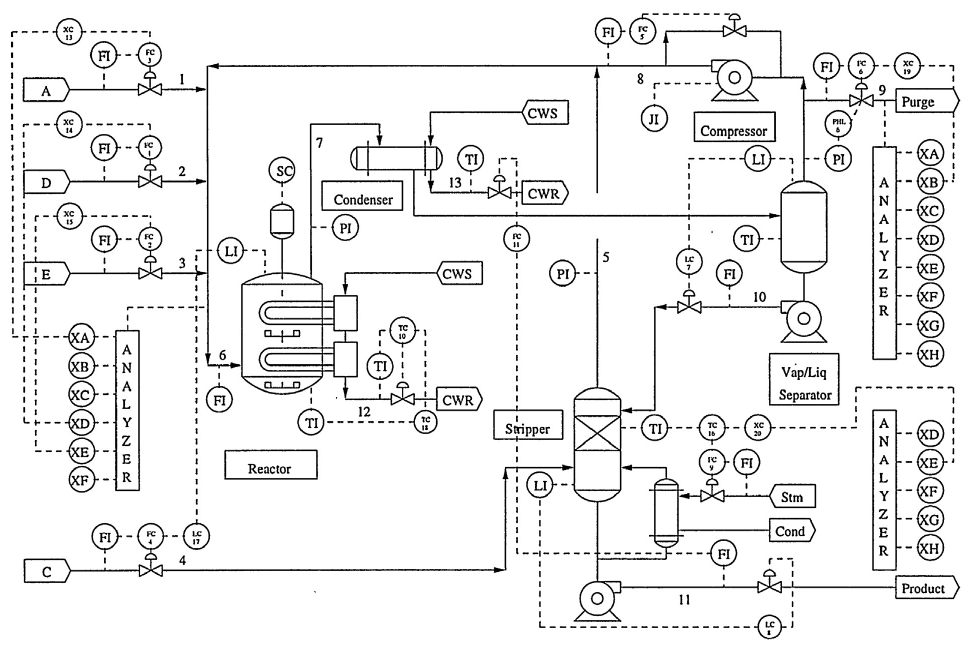}
    \caption{Flow chart of the TE process\cite{LAWRENCERICKER1996205}}
    \label{te_dataset}
\end{figure}

\noindent
\textbf{Dataset.} Based on actual chemical reaction processes, Eastman Chemical Company has developed an open and challenging chemical model simulation platform, the Tennessee Eastman (TE) \cite{LAWRENCERICKER1996205} simulation platform \ref{te_dataset}, which generates time-varying data for testing control and diagnostic models of complex industrial processes. 

The TE dataset contains 21 faults. In process monitoring, the experiment uses normal data as the training set and 21 fault data as test data to evaluate the performance. In the OSFD task, the training set consists of normal data and known faults (fault 1, fault 2, fault 4, and fault 5). Then, the unseen faults (faults 6, 7, 13, 14) are randomly picked and used as test sets to assess the model's performance.

\noindent
\textbf{Training Scheme.} In the ICL-OD method, the model employs Adma Optimizer with a learning rate of 0.001. We set the subvector's length $l=2$ and temperature $\tau=0.01$. The $F$ and $G$ are fully-connected networks. The hidden layers in the $F$ network have $u$ and $2u$ units. A LeakyRelu activation and batch normalization layer is applied for each hidden layer. In the $G$ network, hidden units are $u/4$ and $u/2$, with the LeakyRelu activation for each layer. Unlike network $F$, $G$ applied batch normalization only in the first hidden layer. We set hidden unit $u=200$ for both $F$ and $G$. The classifier $\mathcal{C}$ for the TE dataset is a fully connected network with one hidden layer of 20 units. The classifier applies Adam Optimizer and the CrossEntrophy loss in the training phase. The threshold ${\theta}_{u}$ is set at values when $k$\% = 98\% to reject the unknown samples.

\noindent
\textbf{Results.} In process monitoring, our proposed method outperforms the baseline methods in most cases, highlighted in green in Table \ref{pmTE}. ``OT'' in Table \ref{pmTE} refers to the cases where the event of fault detection was impossible to determine because the indicator value was already over the threshold before the fault introduction. In most cases, our model achieves the highest FDR compared to other baselines. Especially in cases of fault 3, fault 9, and fault 15, our model dramatically improves the results compared to the poor performance of baseline methods. Even in the case where the FDR of our model is not the highest, our proposed method can still achieve relatively good results.

\begin{table*}[ht]
\centering
\caption{Results of process monitoring in TE process}
\begin{tabular}{ccccccccccccccc}
\hline
fault   & Ours           & \multicolumn{2}{c}{PCA\cite{7795255}} & \multicolumn{2}{c}{KPCA\cite{7795255}} & \multicolumn{2}{c}{ICA\cite{GE20131}} & \multicolumn{2}{c}{KICA\cite{KICAresults}} & \multicolumn{2}{c}{DPCA\cite{RATO2013101}} & SJCCA\cite{9068308} & OSA\cite{LOU2022110148}
& DSVDD \cite{wang2023decentralized} \\ \cline{2-15} 
        &        -        & ${T}^{2}$                  & ${Q}$                     & ${T}^{2}$                  & ${Q}$                      & ${I}^{2}$                   & ${SPE}$                  & ${T}^{2}$                    & ${SPE}$                      & ${T}^{2}$                      & ${Q}$                 &    -    & ${SPE}$          &             -                                       \\ \hline
\rowcolor[HTML]{C8FF99} 
1       & \textbf{100}   & 99.50                      & 99.80                     & 99.80                      & 99.80                      & 99.77                       & 99.78                    & 100                          & 100                          & 99.00                          & 99.40                      & 99.75                                & 99.90                                     & 99.75                                              \\
2       & 99.12          & 98.30                      & 98.80                     & 98.80                      & 98.50                      & 98.22                       & 98.27                    & 98.00                        & 98.00                        & 98.40                          & 98.10                      & \textbf{99.47}                       & 95.60                                     & 98.63                                              \\
\rowcolor[HTML]{C8FF99} 
3       & \textbf{61.88} & 8.10                       & 7.80                      & 8.00                       & 7.50                       & 3.46                        & 9.34                     & 6.00                         & 3.00                         & 3.50                           & 1.00                       & 41.38                                & 3.3                                      & 13.25                                             \\
4       & 76.75          & 28.90                      & \textbf{100}              & \textbf{100}               & 37.30                      & 1.96                        & 4.82                     & 82.00                        & \textbf{100}                 & 16.50                          & 99.90                      & \textbf{100}                         & \textbf{100}                             & \textbf{100}                                       \\
5       & 73.75          & 30.60                      & 31.30                     & 28.60                      & \textbf{99.50}             & 22.63                       & 28.51                    & 29.00                        & 27.00                        & 29.30                          & 22.80                      & 36.62                                & 23.00                                       & 33.00                                                 \\
\rowcolor[HTML]{C8FF99} 
6       & \textbf{100}   & 99.30                      & \textbf{100}              & 99.50                      & \textbf{100}               & 99.81                       & 99.96                    & \textbf{100}                 & \textbf{100}                 & 98.90                          & 99.90                      & \textbf{100}                         & \textbf{100}                             & \textbf{100}                                       \\
\rowcolor[HTML]{C8FF99} 
7       & \textbf{100}   & \textbf{100}               & \textbf{100}              & \textbf{100}               & 99.90                      & 36.90                       & 42.52                    & \textbf{100}                 & \textbf{100}                 & 98.60                          & 99.90                      & \textbf{100}                         & \textbf{100}                             & \textbf{100}         \\
\rowcolor[HTML]{C8FF99}
8       & \textbf{99.75} & 97.50                      & 97.90                     & 98.30                      & 98.10                      & 95.85                       & 98.16                    & 97.00                        & 98.00                        & 97.30                          & 97.40                      & 97.85                                & 88.9                                     & 98                                                 \\
\rowcolor[HTML]{C8FF99} 
9       & \textbf{56.12} & 7.40                       & 6.00                      & 6.50                       & 4.50                       & 3.18                        & 8.59                     & 5.00                         & 3.00                         & 3.00                           & 0.20                       & 40.87                                & 1.90                                      & 10.63                                              \\
\rowcolor[HTML]{C8FF99} 
10      & \textbf{87.00} & 49.80                      & 53.90                     & 54.90                      & 86.90                      & 57.70                       & 68.15                    & 81.00                        & 80.00                        & 43.90                          & 17.20                      & 39.58                                & 35.1                                     & 59.75                                              \\
11      & 79.50          & 47.40                      & 73.30                     & 79.30                      & 51.80                      & 32.16                       & 39.24                    & 81.00                        & 77.00                        & 34.00                          & 82.90                      & 78.50                                & 76.5                                     & 73.88                                              \\
\rowcolor[HTML]{C8FF99} 
12      & \textbf{99.50} & 99.00                      & 97.80                     & 99.10                      & 99.50                      & 95.05                       & 99.00                    & 97.00                        & 98.00                        & 99.00                          & 96.40                      & 96.37                                & 89.6                                     & 99.13                                              \\
\rowcolor[HTML]{C8FF99} 
13      & \textbf{98.38} & 95.00                      & 95.40                     & 95.50                      & 95.90                      & 94.16                       & 94.82                    & 95.00                        & 95.00                        & 94.30                          & 95.00                      & 95.29                                & 95.3                                     & 95.00                                                 \\
14      & 95.75          & 99.00                      & 100                       & 100                        & 99.90                      & 99.76                       & 99.92                    & \textbf{100}                 & \textbf{100}                 & 99.00                          & 99.90                      & 89.82                                & \textbf{100}                             & \textbf{100}                                       \\
\rowcolor[HTML]{C8FF99} 
15      & \textbf{52.50} & 12.40                      & 8.80                      & 13.30                      & 14.00                      & 6.49                        & 12.97                    & 5.00                         & 7.00                         & 5.90                           & 0.90                       & 42.37                                & 2.90                                      & 18.50                                               \\
\rowcolor[HTML]{C8FF99} 
16      & \textbf{93.50} & 32.50                      & 48.30                     & 37.00                      & 90.00                      & 24.20                       & 39.89                    & 80.00                        & 52.00                        & 21.70                          & 14.50                      & 26.37                                & 35.10                                     & 52.63                                              \\
17      & 85.75          & 81.60                      & 93.90                     & \textbf{96.10}             & 90.50                      & 88.51                       & 95.38                    & 95.00                        & 95.00                        & 79.00                          & 95.30                      & 41.75                                & 95.9                                     & 91.38                                              \\
\rowcolor[HTML]{C8FF99} 
18      & \textbf{95.88} & 89.50                      & 91.40                     & 91.30                      & 89.40                      & 90.06                       & 90.08                    & 90.00                        & 80.00                        & 89.00                          & 89.80                      & 94.12                                & 90.3                                     & 90.13                                              \\
19      & 65.62          & 8.40                       & 29.10                     & 19.10                      & \textbf{80.80}             & 7.68                        & 22.99                    & 75.00                        & 69.00                        & 4.60                           & 29.80                      & 29.93                                & 18.8                                     & 13.63                                              \\
\rowcolor[HTML]{C8FF99} 
20      & \textbf{84.00} & 47.00                      & 57.30                     & 68.40                      & 72.30                      & 49.79                       & 57.42                    & 58.00                        & 55.00                        & 40.80                          & 49.30                      & 55.75                                & 53.1                                     & 58.63                                              \\
21      & 70.75          & 39.40                      & 51.10                     & 54.50                      & 44.30                      & 38.17                       & 43.74                    & 61.00                        & 58.00                        & 42.90                          & 40.90                      & \textbf{96.90}                       & 57.6                                     & 42.88                                              \\ \hline
Average & \textbf{84.55} & 60.50                      & 68.66                     & 68.95                      & 74.30                      & 54.55                       & 59.69                    & 73.10                        & 71.19                        & 57.08                          & 63.36                      & 71.54                                & 64.90                                    & 68.99                                              \\ \hline
\end{tabular}
\label{pmTE}
\end{table*}

\begin{figure*}[htbp]
    \centering
    \includegraphics[width=18cm]{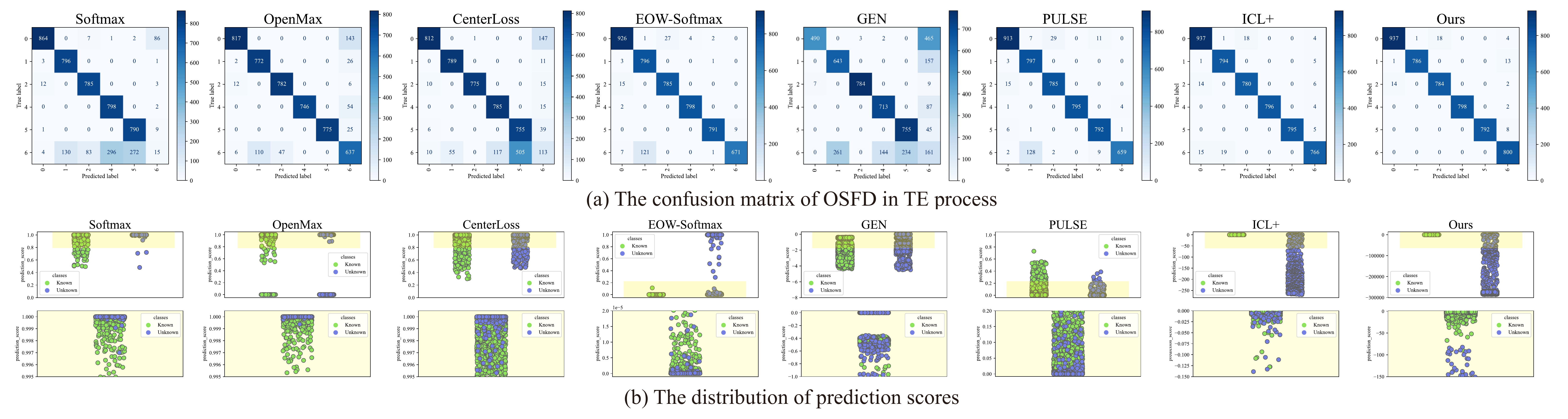}
    \caption{(a) The confusion matrix for the OSFD task in the TE process with Fault 6 as the unknown class. (b) we visualize the distribution of prediction scores generated by each method. The images in the second row are the enlarged display of the yellow region in the first-row images, providing a better view of details. The prediction scores of the Softmax, OpenMax, and CenterLoss methods are softmax probabilities. Scores below the threshold are classified as unknown. The prediction scores of the EOW-Softmax method constitute the (K+1)th-dimensional probability of its output, estimating open-world uncertainty. Scores surpassing the threshold are classified as unknown. 
    The GEN method can be applied to any pre-trained softmax-based classifier to generate the entropy-based score, and scores below a threshold are determined to be unknown. The prediction score of the PULSE method is the output of the discriminator. Scores below the threshold are labeled as unknown classes.
    The prediction score of ICL+ and our method are the opposites of the output score. Data with prediction scores below the threshold are classified as the unknown class. The smaller the overlap between the prediction scores of known and unknown categories, the better the method's performance.}
    \label{te_cm}
\end{figure*}

In OSFD, we repeat experiments for each method five times to report the average F1-score and AUROC. The results in Table \ref{open-set fault diagnosisTE1} show that our proposed model outperforms the baseline methods. Fig. \ref{te_cm} presents the detailed results where Fault 6 is seen as the unknown. From the confusion matrix shown in Fig. \ref{te_cm} (a), we can see that our proposed method has a higher recognition rate for the unknown, while the classification of the closed-set data is still guaranteed.

\begin{table*}[ht]
\centering
\caption{The results of OSFD in TE process}
\begin{tabular}{clcccccccc}
\hline
Unknown fault             & \multicolumn{1}{c}{Metric} & SoftMax     & OpenMax     & CenterLoss  & EOW-Softmax & GEN       & PULSE      & ICL+        & Ours                 \\ \hline
\multirow{2}{*}{Fault 6}  & AUROC                      & .305 ± .090 & .904 ± .057 & .363 ± .102 & .958 ± .027 & .251 ± .112 & .790 ± .276 & .999 ± .001 & \textbf{1.0 ± 0.0}   \\
                          & F1                         & .815 ± .006 & .912 ± .058 & .804 ± .014 & .964 ± .041 & .825 ± .002 & .918 ± .067 & .980 ± .002 & \textbf{.987 ± .003} \\ \hline
\multirow{2}{*}{Fault 7}  & AUROC                      & .858 ± .122 & .897 ± .042 & .865 ± .040 & .744 ± .088 & .788 ± .035 & .925 ± .035 & .995 ± .010 & \textbf{1.0 ± .001}  \\
                          & F1                         & .856 ± .018 & .885 ± .027 & .854 ± .016 & .832 ± .005 & .825 ± .003 & .858 ± .023 & .975 ± .030 & \textbf{.985 ± .003} \\ \hline
\multirow{2}{*}{Fault 13} & AUROC                      & .718 ± .072 & .798 ± .046 & .789 ± .097 & .731 ± .057 & .643 ± .080 & .797 ± .073 & .891 ± .009 & \textbf{.926 ± .005} \\
                          & F1                         & .835 ± .007 & .865 ± .012 & .842 ± .028 & .856 ± .016 & .825 ± .003 & .855 ± .026 & .902 ± .003 & \textbf{.921 ± .011} \\ \hline
\multirow{2}{*}{Fault 14} & AUROC                      & .567 ± .114 & .707 ± .044 & .771 ± .074 & .743 ± .103 & .422± .041  & .844 ± .056 & .815 ± .036 & \textbf{.888 ± .023} \\
                          & F1                         & .828 ± .006 & .868 ± .005 & .836 ± .019 & .862 ± .013 & .828 ± .002 & .870 ± .015 & .914 ± .036 & \textbf{.930 ± .001} \\ \hline
\end{tabular}
\label{open-set fault diagnosisTE1}
\end{table*}

Fig. \ref{te_cm} (b) presents the distribution of the prediction scores to further analyze the superiority of our method. The prediction scores generated by the Softmax and CenterLoss methods exhibit significant overlap between known and unknown classes, leading to the misclassification of most unknown classes as known ones. The OpenMax method demonstrates a clearer distinction between scores of unknown and known classes, but there are still instances where the scores for unknown categories are too high, resulting in misclassification as known categories. The EOW-SoftMax method presents a similar issue, producing overly low scores for unknown data, resulting in misidentification as known classes.
The GEN method has difficulty generating predictive scores with high discrimination, and most of its normal samples (class 0) and unknown samples have similar scores, resulting in poor classification. The PULSE method has better classification results, but some unknown samples have overly high confidence scores, leading to their misclassification into known categories.

\begin{figure}[htbp]
    \centering
    \includegraphics[width=\columnwidth]{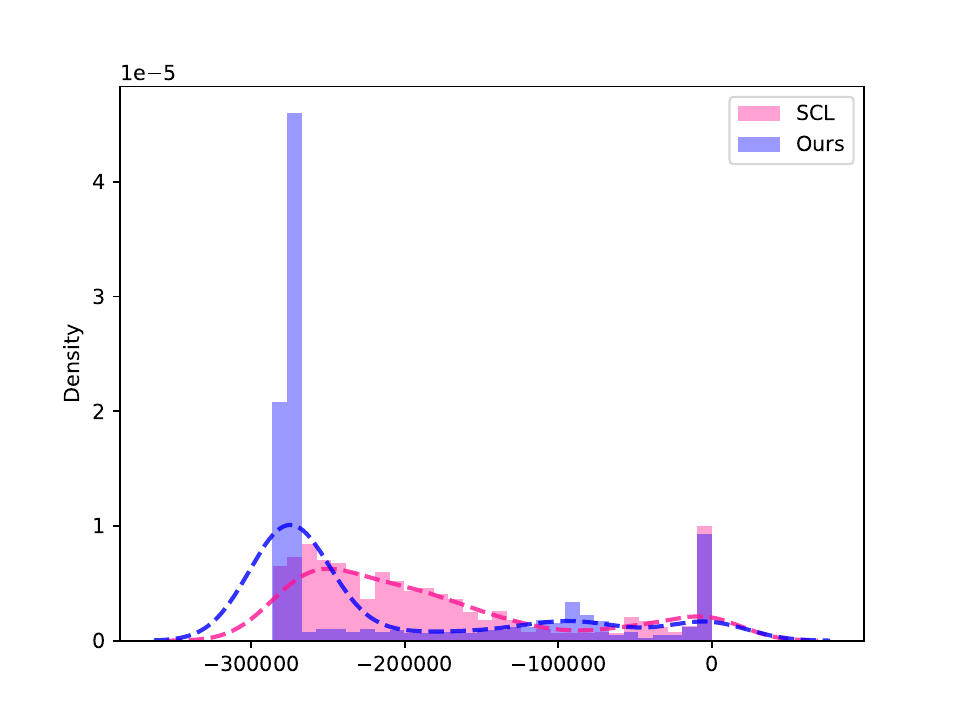}
    \caption{The histogram of prediction scores for unknown classes in ICL+ and our methods. The prediction scores of the ICL+ and our method shown in Fig. \ref{te_cm} (b) are overly dense, making it difficult to discern specific distributional features. Therefore, we present a clearer display of the distribution of the prediction scores for unknown classes.}
    \label{icl-vs-ours}
\end{figure}

However, our method can generate prediction scores with greater distinctiveness, enabling better identification of unknown classes. As shown in Fig. \ref{te_cm} (b), our method has minimal overlap between the distribution of prediction scores for unknown and known classes. In contrast, in the ICL+ method, there is still a significant overlap between the prediction scores of unknown and known classes, resulting in a higher mislabeling rate for unknown classes. Fig. \ref{icl-vs-ours} shows the histogram of the prediction scores for unknown classes in both the ICL+ method and our approach. Our method can generate more ``extreme'' prediction scores for unknown classes, thereby enhancing differentiation from known categories. These results indicate that, in contrast to other baseline methods, our approach with an effective feature extractor can generate prediction scores with greater distribution disparity between unknown and known categories, thereby achieving superior performance.

Experiment I trains and validates the model with the TE dataset and demonstrates that our proposed model can combine the process monitoring and OSFD tasks while achieving better performance than baseline methods in both tasks.

\subsection{Experiment II: evaluation with Multi-phase Flow Facility}

\noindent
\textbf{Datasets.} To further demonstrate the effectiveness of our method, Experiment II uses a real-world dataset TE and the Multiphase Flow Facility (MFF) dataset \cite{MFF} to validate the model. 

The MMF dataset is derived from the Three-phase Flow Facility system at Cranfield University. The data has 24 process variables and is acquired at the sampling rate of 1 Hz. The MFF dataset contains normal data and six fault cases. In normal data, three datasets (T1, T2, T3) are captured from the system to represent normal operating conditions adequately. For each fault case, the process data is generated under different operation conditions and used as different sets. In process monitoring, we use the T1 and T3 sets of normal data to train our method. The description of the test set is shown in Table \ref{testset}. All test datasets are from Set 1 with changing operating conditions. In the OSFD test of Experiment II, we use the T2 set of normal data and Set-2 of each fault case to train and evaluate our model. We randomly select normal data and three fault cases as the training set.

\begin{table}[ht]
\centering
\caption{The description of test sets of MMF in process monitor experiment}
\begin{tabular}{ccc}
\hline
Test Set & Fault Description              & Dataset \\ \hline
1        & Air line blockage              & 1.1     \\
2        & Water line blockage            & 3.1     \\
3        & Top separator input blockage & 4.1     \\
4        & Open direct bypass             & 5.1     \\ \hline
\end{tabular}
\label{testset}
\end{table}

\noindent
\textbf{Training Scheme.} The parameter settings of the ICL-OD method in Experiment II are the same as those in Experiment I. The classifier $\mathcal{C}$ for the MFF dataset is a fully connected network with one hidden layer. The hidden layer has 12 units, followed by a ReLU activation layer. During the training phase, the classifier applies Adam Optimizer and CrossEntrophy loss.

\noindent
\textbf{Results.} The FDR results of the process monitor experiment are shown in Table \ref{mff-pm-result}. It can be seen that our method outperforms all other baseline methods in the MFF dataset.

\begin{table*}
\centering
\caption{The results of process monitoring in MFF}
\begin{tabular}{cccccccccc}
\hline
\multirow{2}{*}{Test Set} & \multicolumn{2}{c}{PCA} & \multicolumn{2}{c}{DPCA} & \multicolumn{2}{c}{PLS} & EDLV          & OARE-GLPP & Ours           \\ \cline{2-10} 
                          & ${T}^{2}$    & ${Q}$    & ${T}^{ 2}$     & ${Q}$    & ${T}^{2}$    & ${Q}$    & ${T}^{2}_{S}$ & -         & -              \\ \hline
1                   & 22.65        & 52.72    & 23.1          & 60.17    & 36.49        & 30.23    & 79.56         & 76.15     & \textbf{100}   \\
2                   & 98.37        & 99.72    & 98.44         & 100      & 99.36        & 98.67    & 99.14         & 98.16     & \textbf{100}   \\
3                   & 34.5         & 92.64    & 35.61         & 94.14    & 40.34        & 43.02    & 91.61         & 66.01     & \textbf{99.42} \\
4                   & 70.63        & OT       & 71.87         & OT       & 94.72        & 58.63    & 89.64         & 42.36     & \textbf{98.97} \\ \hline
\end{tabular}
\label{mff-pm-result}
\end{table*}

For the OSFD evaluation, we ran the experiment five times, and the average results of the MFF process are shown in Table \ref{MMF-results}. The Known-Unknown in Table \ref{MMF-results} shows the known classes for training and their corresponding unknown fault for evaluation in each case. Compared with the other six baseline methods, our approach shows a significant improvement in the F1-score and AUROC metrics for each case. Furthermore, our method outperforms ICL+.
With unknown faults of 2 and 4, both the AUROC and F1-score of our method are significantly better than those of the ICL+ method. Especially when the unknown fault is 2, our method improves by 12\% on the F1-score compared to ICL+, which is a great classification improvement. In the case of unknown faults of 3 and 5, even though AUROC is both 1.0 (achieving the best result), the F1-score of our method still exceeds the ICL+ method, which suggests that our method can better predict the boundaries of known classes and generate more accurate thresholds to reject unknown classes.
The results of Experiment II indicate that even in real-world data, our method can achieve remarkable performance in the GOOFD task and can surpass baseline methods in both tasks. 

\begin{table*}[ht]
\centering
\caption{The results of OSFD in MMF}
\begin{tabular}{clcccccccc}
\hline
Known-Unknown                   & \multicolumn{1}{c}{Metric} & SoftMax     & OpenMax     & CenterLoss  & EOW-SoftMax & GEN         & PULSE       & ICL+        & Ours                 \\ \hline
\multirow{2}{*}{0, 1, 3, 4 - 2} & AUROC                      & .925 ± .036 & .928 ± .059 & .865 ± .146 & .824 ± .127 & .922 ± .017 & .840 ± .173 & .938 ± .038 & \textbf{.960 ± .016} \\
                                & F1                         & .793 ± .000 & .849 ± .045 & .789 ± .075 & .803 ± .129 & .822 ± .103 & .890 ± .105 & .821 ± .084 & \textbf{.941 ± .038} \\ \hline
\multirow{2}{*}{0, 1, 2, 6 - 3} & AUROC                      & .939 ± .070 & .759 ± .093 & .925 ± .046 & .967 ± .027 & .930 ± .056 & .503 ± .252 & 1. ± .000   & \textbf{1. ± .000}   \\
                                & F1                         & .839 ± .060 & .726 ± .029 & .869 ± .051 & .958 ± .018 & .884 ± .054 & .678 ± .080 & .957 ± .082 & \textbf{.998 ± .000} \\ \hline
\multirow{2}{*}{0, 1, 2, 6 - 4} & AUROC                      & .884 ± .199 & .881 ± .052 & .738 ± .126 & .891 ± .070 & .936 ± .041 & .496 ± .028 & .960 ± .072 & \textbf{.998 ± .005} \\
                                & F1                         & .822 ± .040 & .780 ± .063 & .813 ± .031 & .892 ± .046 & .896 ± .041 & .796 ± .033 & .907 ± .072 & \textbf{.969 ± .072} \\ \hline
\multirow{2}{*}{0, 1, 2, 6 - 5} & AUROC                      & .398 ± .094 & .809 ± .041 & .988 ± .012 & .951 ± .030 & .232 ± .052 & .928 ± .063 & 1. ± .000   & \textbf{1. ± .000}   \\
                                & F1                         & .800 ± .006 & .715 ± .053 & .936 ± .052 & .897 ± .061 & .785 ± .026 & .887 ± .055 & .983 ± .033 & \textbf{1. ± .000}   \\ \hline
\end{tabular}
\label{MMF-results}
\end{table*} 

\subsection{Ablation study}
In the ablation study, the experiments are conducted on the TE dataset in the OSFD task. The known class consists of normal data, faults 1, 2, 4, and 5. The experiments are repeated five times to present the average F1-score and AUROC.

\begin{figure}[htbp]
    \centering
    \includegraphics[width=\columnwidth]{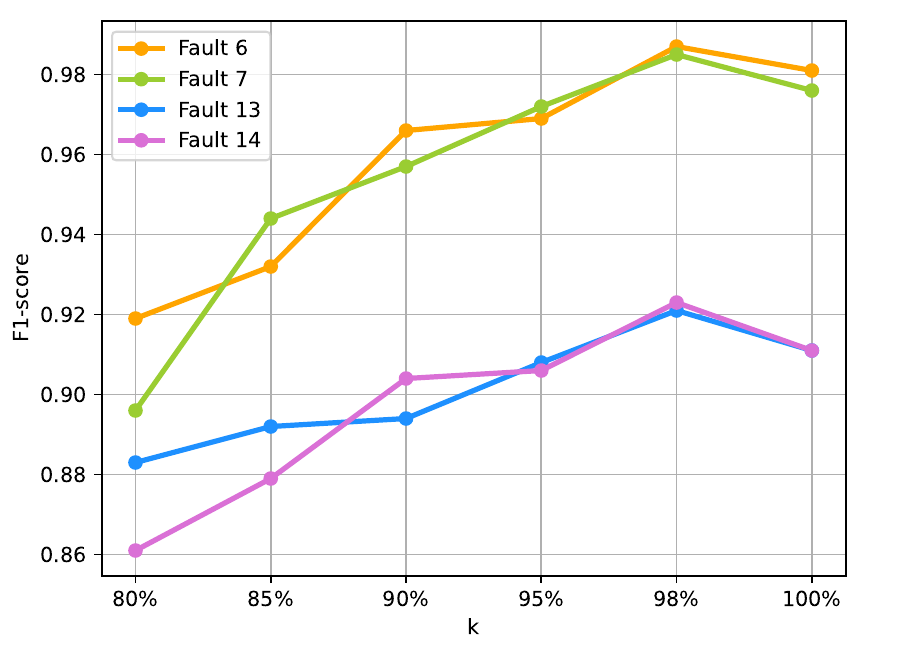}
    \caption{The average F1-score of OSFD tasks in TE process when applying various threshold ${\theta}_{u}$.}
    \label{ablation-threshold}
\end{figure}

\noindent
\textbf{Threshold for outliers.} The threshold ${\theta}_{u}$ is used to determine and reject unknown classes. Therefore, it is crucial to select an appropriate threshold, which will determine the performance of unknown fault detection. We set different values of ${\theta}_{u}$ (80\%, 85\%, 90\%, 95\%, 98\% and 100\%) on the TE dataset in OSFD task. The results in Fig. \ref{ablation-threshold} show that it can reach the best performance when ${\theta}_{u}=98\%$. When the threshold is within the range of 80\% to 98\%, the F1-score shows an increasing trend, indicating that raising the threshold for rejecting unknown categories has a positive effect on the method's performance. However, when the threshold exceeds 98\%, the F1-score decreases, suggesting that setting an overly high threshold for unknown categories may mistakenly identify them as known, leading to a decline in model performance.

\noindent
\textbf{Distance for outlier detection.} In order to find a more suitable measurement for unknown class detection, we compared the Mahalanobis distance with other measures of distance (city-block distance, Canberra distance, and Euclidean distance). The results in Table \ref{ablation-study-distance} show that the Mahalanobis distance leads to better classification performance. 
Both AUROC and F1-score are significantly better than the rest of the distances in the case of the Mahalanobis distance. Particularly, in the case of the unknown fault 13, the AUROC of the Mahalanobis distance is 12.6\%-22\% higher compared to the rest of the distances. This suggests that the Mahalanobis distance generates more discriminatory confidence scores compared to the other distances.
Theoretically, the Mahalanobis distance can be regarded as a modification of the Euclidean distance, which corrects the problem of inconsistent and correlated dimensional scales, as shown in \eqref{md}. Therefore, since the Mahalanobis distance can eliminate the problem of different scales between different dimensions, and make dimensional corrections to the sample distribution, it is more suitable for outlier detection than other distances.

\begin{table*}
\centering
\caption{Ablation study on different measurements of distance}
\begin{tabular}{ccccccccc}
\hline
Unknown     & \multicolumn{2}{c}{Fault 6}             & \multicolumn{2}{c}{Fault 7}                & \multicolumn{2}{c}{Fault 13}                & \multicolumn{2}{c}{Fault 14}                \\ \hline
Metirc      & AUROC            & F1                   & AUROC               & F1                   & AUROC                & F1                   & AUROC                & F1                   \\ \hline
City-block  & .977 ± .012      & .919 ± .064          & .994 ± .013         & .961 ± .061         & .785 ± .122          & .883 ± .030          & .782 ± .069          & .814 ± .033          \\
Euclidean   & .999 ± .001      & .976 ± .019          & .992 ± .019         & .960 ± .046          & .756 ± .092          & .880 ± .038          & .775 ± .134          & .895 ± .048          \\
Canberra    & .984 ± .010      & .973 ± .009          & .907 ± .163         & .928 ± .075          & .691 ± .120          & .870 ± .024          & .790 ± .057          & .869 ± .038          \\
Mahalanobis & \textbf{1.0 ± 0} & \textbf{.987 ± .003} & \textbf{1.0 ± .001} & \textbf{.985 ± .003} & \textbf{.911 ± .017} & \textbf{.928 ± .007} & \textbf{.803 ± .081} & \textbf{.908 ± .026} \\ \hline
\end{tabular}
\label{ablation-study-distance}
\end{table*}

\section{Conclusion}
In this paper, we are the first to introduce the novel idea of an integrated fault diagnosis framework and present the GOOFD framework to integrate multiple fault diagnosis tasks.
Compared to existing single-task fault diagnosis approaches, our ICL-OD method can be applied to multiple  tasks and can also better extract features and generate prediction scores with greater discriminability.
Extensive experiments are conducted to demonstrate that our method leads to better performance for each sub-task. 
In future work, we intend to expand GOOFD to cover more diagnostic subtasks, enhancing fault detection capabilities and system robustness. This development aims to improve industrial safety, and machine reliability, and reduce maintenance costs. Furthermore, we will work on incorporating interpretability within the framework to improve transparency and user trust, which are expected to deliver significant societal advantages, such as safer industrial operations and economic efficiencies.

\footnotesize{
% \bibliographystyle{unsrt}
% \bibliographystyle{unsrtnat}%
% \bibliography{IEEEabrv,ref/bibfile}
\bibliographystyle{IEEEtran}
\bibliography{ref.bib}
}

\vspace{12pt}

\begin{IEEEbiography}[{\includegraphics[width=1in,height=1.25in,clip,keepaspectratio]{img/Xinyue_Wang.jpeg}}]{Xinyue Wang} is currently a graduate student at the Zhejiang University—University of Illinois at Urbana-Champaign Joint Institute, Zhejiang University, Haining, China. Also, she received her undergraduate degree from the Beijing University of Posts and Telecommunications in the major of Digital Media Technology in 2022. Her field interests are artificial intelligence, natural language processing, and fault diagnosis.
\end{IEEEbiography}

\begin{IEEEbiography}[{\includegraphics[width=1in,height=1.25in,clip,keepaspectratio]{img/Hanrong_Zhang.jpeg}}]{Hanrong Zhang} received a dual B.S. degree in Computer Science from the University of Leeds, Leeds, United Kingdom, and Southwest Jiaotong University, Chengdu, China, in 2022. He is currently pursuing a Master degree with the Zhejiang University—University of Illinois at Urbana-Champaign Joint Institute, Zhejiang University, Haining, China. His current research interests include deep learning, fault diagnosis, and knowledge graph. 
\end{IEEEbiography}

\begin{IEEEbiography}[{\includegraphics[width=1in,height=1.25in,clip,keepaspectratio]{img/Xinlong_Qiao.jpeg}}]{Xinlong Qiao} received the B.E. degree in software engineering from the Harbin Institute of Technology, Weihai, China, in 2023. He is currently pursuing a Master's degree at the Zhejiang University-University of Illinois at Urbana-Champaign Joint Institute, Zhejiang University, Haining, China. His current research interests include deep learning, named entity recognition, and fault diagnosis.
\end{IEEEbiography}

\begin{IEEEbiography}[{\includegraphics[width=1in,height=1.25in,clip,keepaspectratio]{img/Ke_Ma.jpeg}}]{Ke Ma} is currently a doctoral student in Computer Science at Zhejiang University. He received his M.S. degree in Structural Engineering, Mechanics and Materials from Department of Civil and Environmental Engineering, University of California, Berkeley in 2020.
\end{IEEEbiography}

\begin{IEEEbiography}[{\includegraphics[width=1in,height=1.25in,clip,keepaspectratio]{img/Shuting_Tao.jpeg}}]{Shuting Tao} is currently a doctoral student at the College of Computer Science and Technology in Zhejiang University, China. She received dual Bachelor degree in Computer Engineering from Zhejiang University and University of Illinois at Urbana-Champaign in 2021. Her current research interests include machine learning, knowledge graph, and imbalanced learning. 
\end{IEEEbiography}

\begin{IEEEbiography}[{\includegraphics[width=1in,height=1.25in,clip,keepaspectratio]{img/Peng_Peng.jpeg}}]{Peng Peng} is currently a doctoral student at the National Engineering Research Centre of Computer Integrated Manufacturing System (CIMS-ERC) in Tsinghua University, Beijing, China.  He received his Bachelor degree at the Department of Automation from Northeastern University in 2016. His research interests are process monitoring and prognostic and health management. 
\end{IEEEbiography}

\begin{IEEEbiography}[{\includegraphics[width=1in,height=1.25in, clip,keepaspectratio]{img/Hongwei_Wang.jpeg}}]{Hongwei Wang} received the B.S. degree in information technology and instrumentation from Zhejiang University, China, in 2004, the M.S. degree in control science and engineering from Tsinghua University, China, in 2007, and the Ph.D. degree in design knowledge retrieval from the University of Cambridge. From 2011 to 2018, he was a Lecturer and then, a Senior Lecturer in engineering design with the University of Portsmouth. He is currently a Tenured Professor with Zhejiang University and the University of Illinois at Urbana–Champaign Joint Institute. His research interests include knowledge engineering, industrial knowledge graph, intelligent and collaborative systems, and data-driven fault diagnosis. His research in these areas has been published in over 120 peer-reviewed papers in well-established journals and international conferences. 
\end{IEEEbiography}